\documentclass[lettersize,journal]{IEEEtran}
\usepackage{amsmath,amsfonts}
\usepackage{algorithmic}
\usepackage{algorithm}
\usepackage{array}
\usepackage[caption=false,font=normalsize,labelfont=sf,textfont=sf]{subfig}
\usepackage{textcomp}
\usepackage{stfloats}
\usepackage{url}
\usepackage{verbatim}
\usepackage{graphicx}
\usepackage{cite}
\usepackage{multirow}
\usepackage{xcolor}
\usepackage{xspace}
\newcommand{\ie}{{\emph{i.e.}}\xspace}
\begin{document}

\title{Beyond Appearance: A Multi-cue Framework and Large-scale Benchmark for Pedestrian Association and Tracking on Mobile Aerial--Ground Platforms}

\author{
Ruiqi Wu\IEEEauthorrefmark{1}, 
Bingliang Jiao\IEEEauthorrefmark{1},
Ruize Han, Hangzheng Yu, Xunkai Jiang, Shining Wang, Yuanqi Hu, 
Wenxuan Wang,
Peng Wang\IEEEauthorrefmark{2}

\thanks{
\IEEEauthorrefmark{1} Equal Contribution. 
\IEEEauthorrefmark{2} Corresponding author. 

Manuscript received April 4, 2026. Ruiqi Wu, Bingliang Jiao, Hangzheng Yu, Xunkai Jiang, Shining Wang, Yuanqi Hu, Wenxuan Wang, and Peng Wang are with the 
School of Computer Science, Northwestern Polytechnical University, Xi'an, China;
Ningbo Institute, Northwestern Polytechnical University, Ningbo 315000, China; 
and National Engineering Laboratory for Integrated Aero-Space-Ground-Ocean Big Data Application Technology, Xi'an, China
(email: wurq@mail.nwpu.edu.cn; wxwang, peng.wang@nwpu.edu.cn).
% Wenxuan Wang is with the 
% School of Computer Science, Northwestern Polytechnical University, Xi'an, China;
% Shenzhen Research Institute of Northwestern Polytechnical University, Shenzhen, China; 
% and National Engineering Laboratory for Integrated Aero-Space-Ground-Ocean Big Data Application Technology, Xi'an, China
% (email:wxwang@nwpu.edu.cn).
Ruize Han is with the Shenzhen University of Advanced Technology, Shenzhen, China
(email: hanruize@suat-sz.edu.cn).
}
}


\markboth{IEEE Transactions on X,~Vol.~X, No.~X, ~2026}%
{Shell \MakeLowercase{\textit{et al.}}: A Sample Article Using IEEEtran.cls for IEEE Journals}

\IEEEpubid{0000--0000/00\$00.00~\copyright~2021 IEEE}

\maketitle

\begin{abstract}
% Multi-view Multi-object Association and Tracking (MvMoAT) seeks to associate target objects across camera views and track them over time, which is the core of multi-platform cooperative perception. 
Multi-view Multi-object Association and Tracking (MvMoAT) is an important task for multi-camera surveillance in information forensics and security, aiming to associate objects across views and track them over time for reliable identity persistence and forensic trajectory reconstruction in multi-platform cooperative perception.
Unlike conventional multiple object tracking, MvMoAT must handle frequent viewpoint shifts that distort target appearance, undermining both cross-view association and temporal tracking. 
To tackle these challenges, we propose a viewpoint-robust MvMoAT framework named Feature Unification framework for multi-view aSsociation and IdentificatiON (FUSION). Within FUSION, the Multi-cue Adaptive Combination (MAC) module adaptively fuses multiple viewpoint-invariant cues with appearance features, mitigating the interference caused by appearance distortion and strengthening cross-view association. Furthermore, the Online Multi-view Feature Synchronization (OMFS) module dynamically fuses pedestrian features across historical and cross-view frames, forming robust trajectory representations for temporally consistent tracking.
In addition, we further introduce a large-scale dataset, termed RealMvMoAT, which exhibits frequent and substantial viewpoint variations both across cameras and within individual cameras, providing a valuable benchmark for evaluating model robustness to viewpoint variation. RealMvMoAT contains $504.9$K frames captured by $7$ cameras ($5$ UAV views and $2$ ground views) across $10$ scenes, with over $7.3$M identity-labeled bounding boxes annotated. All camera views in RealMvMoAT exhibit random and substantial motion. To our best knowledge, RealMvMoAT is \textbf{the largest} MvMoAT dataset to date. 
With its large-scale data volume, diverse viewpoints, complex platform motions, and realistic target trajectories, it could serve as a comprehensive resource for advancing future research in this area. 
% With its large-scale data volume, diverse viewpoints, complex platform motions, and realistic target trajectories, it could serve as a comprehensive resource for advancing future research in surveillance and security-oriented multi-camera video analysis.
Extensive experiments on RealMvMoAT and six public benchmarks demonstrate that FUSION achieves state-of-the-art performance. Our project is available at \textcolor{blue}{\textit{\textbf{https://wurqjackey.github.io/FUSION}}}.

% Additionally, existing datasets, while valuable, often fall short in capturing the full range of viewpoint diversity, dynamic motion, and scale necessary for comprehensive evaluation in real-world scenarios. To address these limitations, we introduce RealMvMoAT, a large-scale benchmark comprising $504.9$K frames from $7$ cameras across $10$ scenes, annotated with over $7.3$M identity-labeled boxes. To our knowledge, it is \textbf{the largest} MvMoAT dataset to date, uniquely offering diverse viewpoints, complex platform motions, and realistic target trajectories, making it a more comprehensive resource for advancing research in this area.
% Finally, extensive experiments on RealMvMoAT and six public benchmarks demonstrate the state-of-the-art performance of FUSION. The dataset and code will be released.
\end{abstract}   

\begin{IEEEkeywords}
Multi-view multi-object tracking, Multi-view association, multi-view tracking.
\end{IEEEkeywords}

\section{Introduction}

\begin{figure}[htbp] 
	\centering
	% \vspace{-5mm}
 \includegraphics[width=1.0\linewidth,scale=1.0]{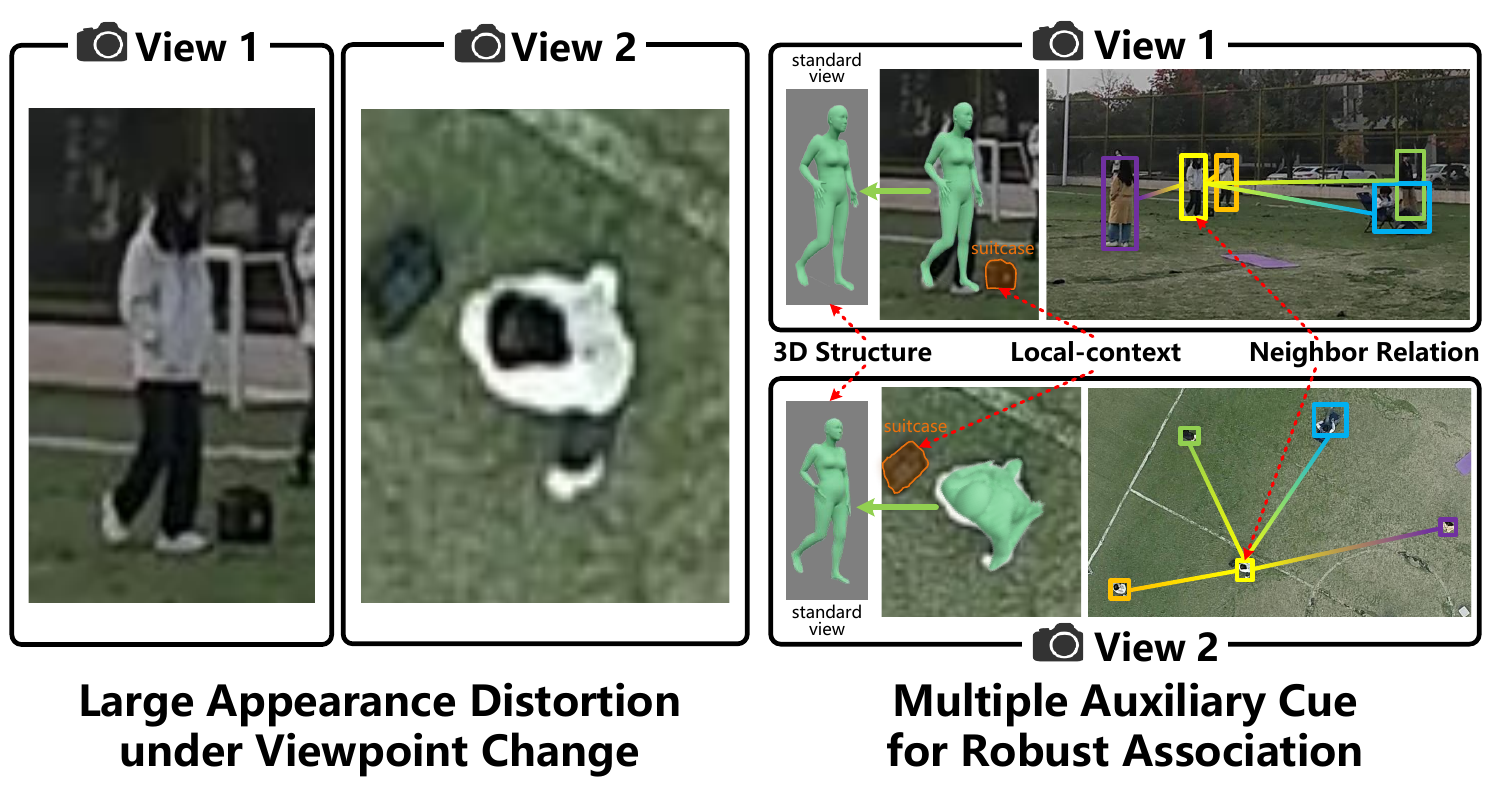}
 \vspace{-7mm}
	\caption{
Under large viewpoint changes \textbf{(left)}, the same person may exhibit drastic appearance differences across cameras, making appearance-based matching unstable and unreliable.
By incorporating auxiliary cues such as local context, 3D structure, and neighbor relations \textbf{(right)}, our FUSION framework can leverage consistent environmental, human-structural, and spatial relational cues to achieve more reliable cross-view identity association.
}
	\label{fig:motivation}
\vspace{-6mm}
\end{figure}

% In modern surveillance and intelligent monitoring systems, individuals are often captured by multiple distributed cameras with overlapping or complementary viewpoints. 
% Maintaining identity consistency across these cameras is essential for reliable person-centric analysis, long-term trajectory reconstruction, and cross-camera event understanding. 
In modern information forensics and security applications, especially surveillance and intelligent monitoring systems, individuals are often captured by multiple distributed cameras with overlapping or complementary viewpoints. 
Maintaining identity consistency across these cameras is essential for reliable security-oriented person-centric analysis, forensic trajectory reconstruction, and cross-camera event understanding.
However, in practical deployments such as aerial--ground collaborative monitoring, dynamic camera motion, and drastic viewpoint changes introduce severe appearance variations, significantly increasing the difficulty of preserving identity continuity across views and over time. To address these challenges, the Multi-view Multi-object Association and Tracking (MvMoAT) task has been proposed to associate identical objects across camera views while maintaining their temporal trajectories. 
With the rapid advancement of UAV technology, aerial--ground collaborative perception has further expanded the practical relevance of this problem in surveillance and public-safety scenarios, where mobile platforms and heterogeneous viewpoints become common.
In this work, we focus on the MvMoAT problem under mobile aerial--ground settings.

The MvMoAT task requires jointly resolving cross-view identity association and temporal tracking, where viewpoint variations often induce significant appearance discrepancies and identity ambiguity. 
Such cross-camera misassociation may fragment identity trajectories and compromise the reliability of surveillance evidence and forensic analysis in multi-camera security systems.
\IEEEpubidadjcol Existing studies on person re-identification~\cite{wang2025secap,jiao2025generalizable,lu2025posture} have explored invariant representation learning for identity matching, while conventional Multiple Object Tracking (MOT)~\cite{chu2019famnet,sun2019deep,tian2021robust,kumar2020p} methods emphasize temporal consistency within individual views. 
However, under mobile multi-camera scenarios with dynamic viewpoint shifts, jointly addressing robust cross-view identity association and long-term trajectory integrity remains challenging. 
Most existing MvMoAT methods~\cite {gan2021self, feng2025Unveiling, hao2024divotrack, han2024benchmarking, liu2023robust} still rely predominantly on pedestrian appearance features for identity matching. 
In real-world mobile aerial-ground deployments, however, drastic viewpoint changes and frequent occlusions often cause severe appearance distortions. As a result, pedestrian matching based solely on appearance cues becomes unreliable.

As shown in Fig.~\ref{fig:motivation}, significant appearance distortions caused by viewpoint differences make appearance-based matching highly challenging.
Recognizing that such discrepancies render existing appearance-based algorithms suboptimal, we draw inspiration from human perception, where identity recognition naturally integrates multiple auxiliary cues beyond mere appearance.
Intuitively, when humans compare and align the same pedestrian across images simultaneously captured from different views, they rely not only on appearance information but also on cues like pedestrians' surrounding environmental context, the pedestrians' intrinsic 3D attributes such as body shape, and their spatial relationships with nearby pedestrians to inform their judgment~\cite{hayes2010neural,rice2013unaware,foster2021separated,johansson1973visual,yan2020learning}. Based on this insight, we propose to comprehensively integrate multiple auxiliary cues with pedestrians' appearance features to achieve robust cross-camera association and tracking.

Practically, we propose \textbf{FUSION} (Feature Unification framework for multi-view aSsociation and IdentificatiON), a novel and robust MvMoAT model. At its core, FUSION contains the Multi-cue Adaptive Combination (MAC)  module, comprising four branches, namely, an appearance branch encoding basic appearance features, a local-context branch capturing environmental cues from the pedestrian’s surroundings, a self-supervised 3D information branch extracting intrinsic 3D attributes, and a neighbor branch leveraging graph neural networks to model spatial inter-object relations. Within MAC, an adaptive fusion mechanism integrates these heterogeneous cues, facilitating robust cross-view association.
Additionally, we design the OMFS (Online Multi-view Feature Synchronization) module, which dynamically fuses each pedestrian’s informative features across historical frames and diverse camera views to construct robust trajectory representations, facilitating precise tracking through accurate matching between historical trajectories and newly detected targets. 

{\color{black} Additionally, to underpin future studies on evaluating model robustness to viewpoint variations, we construct a large-scale real-world dataset named RealMvMoAT, which features frequent and substantial viewpoint variations.
}
The RealMvMoAT comprises $504,900$ high-resolution frames captured by $7$ synchronized cameras across $10$ scenes, totaling over $7,303,191$ annotated bounding boxes, a scale exceeding existing MvMoAT benchmarks by more than five times. Importantly, RealMvMoAT is designed to faithfully mirror the challenges of mobile perception by incorporating complex platform motions and a diverse range of viewpoints, including top-down, oblique aerial, and ground-level perspectives. 
This dataset is highly promising to provide a valuable foundation for future research on real-world MvMoAT tasks.

Extensive experiments on our designed RealMvMoAT dataset and six existing benchmarks demonstrate the effectiveness of FUSION for the MvMoAT task. In summary, the main contributions of this paper include: 
\begin{itemize}

\item We propose \textbf{FUSION}, a novel framework that integrates multiple viewpoint-invariant cues through the \textbf{MAC} module for robust cross-view association and tracking. Besides, a \textbf{OMFS} module is designed to form informative trajectory representations for effective tracking.
\item We introduce \textbf{RealMvMoAT}, a large-scale real-world benchmark with detailed annotations, dynamic camera motion, and diverse viewpoints.
\item Extensive experiments on RealMvMoAT and six other public datasets demonstrate the state-of-the-art performance of our FUSION model.
\end{itemize}

\section{Related Work}

\subsection{Person Re-Identification}

Person re-identification (ReID) aims to retrieve images of same person across different cameras, and has been widely studied due to its importance in intelligent surveillance and person-centric analysis~\cite{wu20193d,luo2020atrong,sun2018beyong,wang2018learning,dong2014deepmetric,Hermans2017indensetriplet,chen2017beyongdquaruplet,qian2020long,jiao2022dynamically,jiao2025generalizable,wu2025enhancing,wang2025secap}.
Related retrieval tasks have also been explored for other object categories, such as vehicle ReID~\cite{WANG2019Vehicle,liu2016large} and animal ReID~\cite{jiao2023towards}.

Existing ReID methods mainly focus on two aspects: feature representation learning and metric learning. 
For feature learning, global-based methods, such as VLAD~\cite{wu20193d} and BNNeck~\cite{luo2020atrong}, learn holistic identity representations from the entire pedestrian image. 
To capture finer-grained cues, part-based methods, e.g., PCB-RPP~\cite{sun2018beyong}, further exploit local body regions to enhance discrimination. 
Some approaches combine global and local representations to leverage their complementary advantages. 
For example, Wang et al.~\cite{wang2018learning} proposed a multiple granularity network that jointly learns one global branch and multiple local branches for more robust person representation. 
In parallel, metric learning methods~\cite{dong2014deepmetric,Hermans2017indensetriplet,chen2017beyongdquaruplet} optimize the embedding space by enlarging inter-identity differences while reducing intra-identity variations, where triplet loss~\cite{Hermans2017indensetriplet} and quadruplet loss~\cite{chen2017beyongdquaruplet} are widely adopted.

Although these ReID methods provide strong foundations for identity matching, most of them mainly rely on appearance representations. 
Under dynamic multi-view scenarios, especially when pedestrians are observed from aerial and ground cameras with large viewpoint changes, the same person may exhibit severe appearance distortion across views, which weakens the reliability of appearance-only matching. 
Therefore, beyond conventional ReID features, it is desirable to introduce more viewpoint-robust cues to support reliable identity association in challenging multi-view settings.

\subsection{MOT and MvMoAT}
Multiple object tracking (MOT) is a fundamental vision task that involves associating multiple targets across frames. Traditional approaches include Tracking-by-Detection (TBD)~\cite{bewley2016simple,du2023strongsort}, Joint Detection and Embedding (JDE)~\cite{wang2021multiple,jin2023multi}, and Joint Detection and Tracking (JDT)~\cite{zhou2020tracking,lu2020retinatrack}. While MOT methods have evolved to handle occlusion, motion, and real-time demands—even in drone~\cite{liu2023robust} or 3D scenes~\cite{weng20203d}—they are primarily designed for monocular, single-view inputs and do not account for cross-view identity alignment.

MvMoAT extends MOT to multi-camera scenarios, where the system must jointly perform cross-view identity association and temporal tracking across synchronized views.
Early approaches~\cite{han2020complementary,liu2023robust,wu2025temporal} focus on two-camera settings. For instance, MIA-Net~\cite{liu2023robust} uses inter-view transformations, and TSMMT~\cite{wu2025temporal} leverages historical trajectories. However, these methods are limited in scalability and often fail in more complex multi-view configurations.
Recent works~\cite{gan2021self,han2022multiview,han2024benchmarking,feng2025Unveiling,hao2024divotrack} extend to three or more views, employing self-supervised constraints such as symmetric and transitive consistency~\cite{gan2021self,feng2025Unveiling}, or using spatial priors and aerial views~\cite{han2024benchmarking}. 
More recently, Self-MVA~\cite{chen2025learning} introduces a novel direction by learning view registration without camera parameters, achieving strong cross-view association in controlled settings. However, it is restricted to static platforms and to viewpoint relations observed during training, failing under dynamic or unseen configurations. 
Overall, most existing approaches still assume stationary cameras and strong scene priors, limiting their effectiveness in dynamic moving camera views.

In contrast, our method tackles multi-view tracking with a multi-cue framework for robust cross-view association and a synchronization module for temporally consistent tracking. By adaptively fusing appearance, local context, 3D structure, and neighbor relations, it improves robustness to frequent viewpoint changes and camera motion.

% We also note that some multi-view multi-human 3D pose estimation and tracking (MVMHPET) methods like EasyRet3D~\cite{yin2025easyret3d} share conceptual similarities with our approach. However, their primary goal is 3D mesh reconstruction and camera pose estimation across views. In contrast, we focus on identity-level association and temporal tracking under dynamic multi-view conditions, rather than high-precision 3D reconstruction. Given these differences in objectives and settings, direct comparisons are not included.

\subsection{MvMoAT Benchmarks}

Several benchmarks have been proposed for MvMoAT research. Early datasets such as Campus~\cite{xu2016multi} and EPFL~\cite{fleuret2008multicamera} use 3--4 fixed cameras to capture synchronized indoor and outdoor scenes, including classrooms, basketball courts, and campus walkways. MvMHAT~\cite{feng2025Unveiling} extends them with limited camera motion and slightly more dynamic viewpoints, while CvMHAT-R~\cite{han2024benchmarking} adds synthetic aerial--ground camera pairs to broaden altitude and perspective coverage. DIVOTrack~\cite{hao2024divotrack} scales up the data and includes surveillance views, but still provides only three cameras per scene with limited motion dynamics. MDMT~\cite{liu2023robust}, a recent large-scale dataset, offers diverse aerial views but remains limited to pairwise associations and stationary drones.

To address these limitations, we introduce a new benchmark featuring more complex camera motion, a broader viewpoint range, and a larger scale to reflect real-world conditions better.

\begin{figure*}[t] 
	\centering
	\includegraphics[width=0.85\linewidth,scale=0.875]{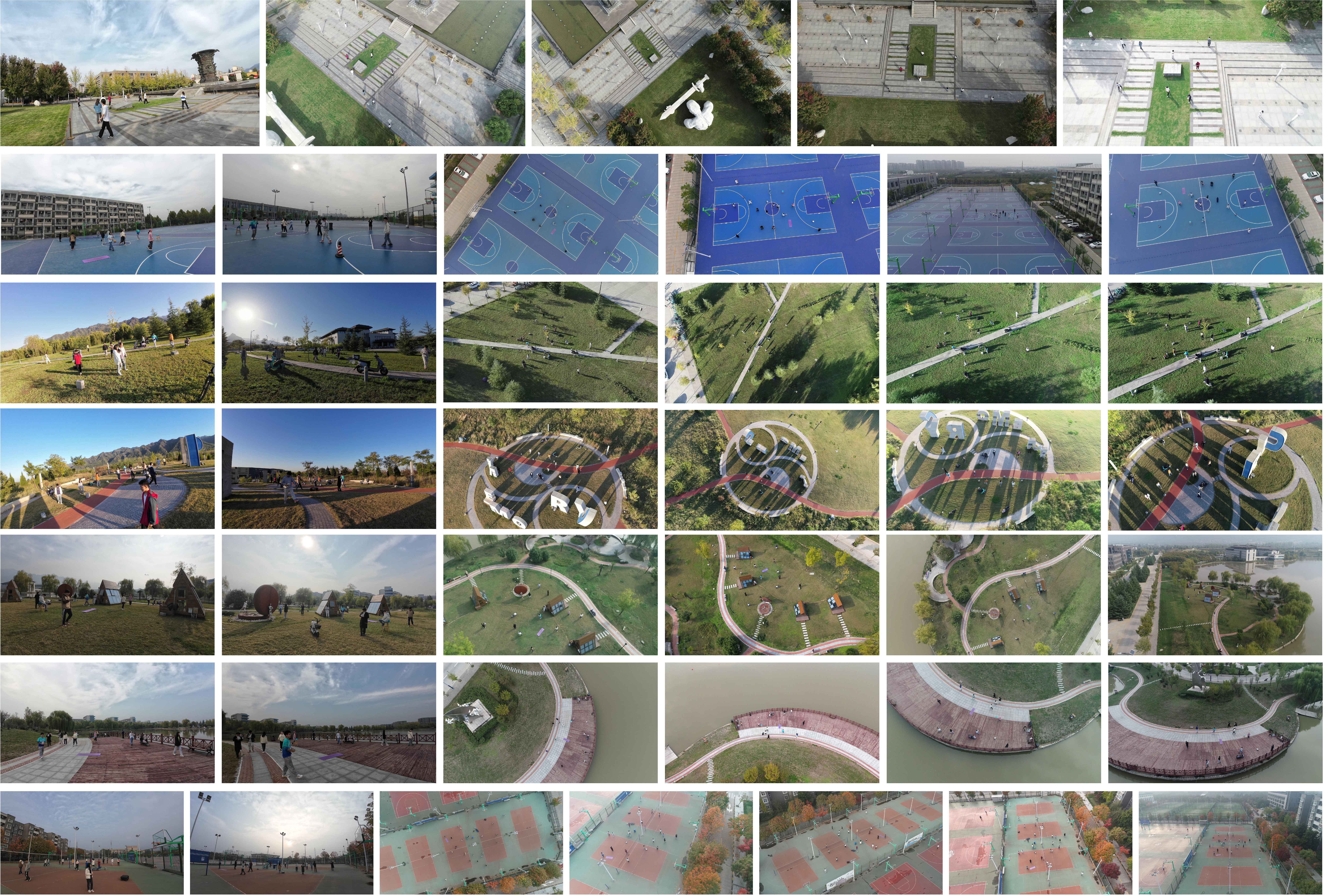}
    \vspace{-2mm}
	\caption{Visualization of the RealMvMoAT benchmark. Each row shows synchronized frames from different viewpoints within the same scene.}
	\label{fig:realmvmoatexp}
\vspace{-2mm}
\end{figure*}

\section{RealMvMoAT Benchmark}
Research on MvMoAT has grown steadily, but progress remains constrained by the limitations of existing public datasets. 
{\color{black} Specifically, existing works are constrained by fixed cameras~\cite{chavdarova2018wildtrack, xu2016multi, fleuret2008multicamera}, a limited number of camera views~\cite{liu2023robust,han2024benchmarking}, or restricted camera motion~\cite{liu2023robust,hao2024divotrack,feng2025Unveiling}, making it difficult to fully capture the impact of complex real-world viewpoint variations on the MvMoAT task. To address this limitation, we propose the RealMvMoAT dataset, which includes a large number of camera views with significant and irregular motion, comprising 5 UAV views and 2 ground views, and exhibiting rich and drastic viewpoint variations.}
% Most benchmarks rely on either static or typically correlated cameras that exhibit predictable motion patterns, often covering only narrow ranges of viewpoints. These datasets also lack sufficient diversity in scene types and real-world dynamics, which limits the development of algorithms intended for deployment in more complex environments.
%To address these limitations, we introduce \textbf{RealMvMoAT}, a large-scale benchmark collected from both aerial and ground mobile platforms. 
% This dataset is designed to reflect actual mobile multi-view perception applications, support both association and tracking tasks, and facilitate research on generalizable and robust multi-view perception systems.
This dataset reflects real-world mobile multi-view perception applications, supporting both association and tracking tasks, and facilitating research on robust, generalizable multi-view systems.

\subsection{Data Collection}

As illustrated in Fig.~\ref{fig:realmvmoatexp}, we select diverse outdoor environments, including parks, lawns, lakesides, basketball courts, football fields, and urban plazas, covering varied spatial layouts and crowd behaviors. Each scene includes one to two ground-view action cameras and up to five drones at different heights and orientations, producing 5--7 shifting viewpoints through motion patterns such as circular sweeps, hovering rotations, lateral panning, and zigzag flight. 
During recording, participants move naturally without predefined scripts, sometimes gathering and sometimes dispersing, simulating real crowd dynamics in surveillance, event monitoring, and public-space analytics. As a result, RealMvMoAT captures authentic variations in movement, interaction, and cross-view observation. 
Each scene is recorded multiple times, yielding 52 groups of synchronized multi-view videos, each with 10--28 pedestrian identities and 5--7 views. In total, the benchmark contains 314 videos and 504{,}900 frames at 30 FPS, all with a resolution of $3200 \times 1800$. 
Data collection required over \textbf{350 man-hours} for scene preparation, actor participation, UAV operation, and synchronized multi-view acquisition.

\begin{table*}[!ht]
% \renewcommand\arraystretch{1.0}
% \large
\begin{center}
\caption{Comparison of existing MvMoAT benchmarks on key aspects such as the number of video groups and videos, total frame count, frame rate, number of cameras, resolution, camera motion type, and viewpoint coverage. Camera motion types include: Static, Light Moving (gentle motion like slight tilting), Regular Moving (steady and moderate motion like consistent tracking), and Irregular Moving (dynamic and unpredictable motion like circular sweeps, hovering rotations, and zigzag flights). 
% RealMvMoAT stands out for its scale and diversity, providing the most comprehensive viewpoints under complex and dynamic camera motion.
}
\vspace{-2mm}
\label{tab:comp_benchmark}

\footnotesize
\renewcommand\arraystretch{0.90}
\setlength{\tabcolsep}{4.1pt}
% \begin{tabular*}{\textwidth}{ l | c c | c c c | c c c c c}

\begin{tabular*}{\textwidth}{l | c c c c c c c c c}
%\hline
% \noalign{\hrule height 1.5pt}%加粗
\hline
Benchmark & \#Group & \#Video & \#Frame & FPS & \#Camera & \#GT-Labels & Resolution & Camera Motion Type & Viewpoint coverage \\
\hline
% MOT17~\cite{sun2019deep} & - &14  & 33,705 & 7–30 & 1 & 901,119 & $1920\times1080$ & No & Ground \\
MvMHAT (EPFL)~\cite{fleuret2008multicamera,feng2025Unveiling} & 8 & 30 & 69,964 & 25 & 3-4 & 455,979 & $360\times288$ & Static & Ground \\
MvMHAT (Campus)~\cite{feng2025Unveiling,xu2016multi} & 6 &22  & 55,705 & \textbf{30} & 3-4 & 311,537 & $1920\times1080$ & Static & Ground \\
CvMHAT-R~\cite{han2024benchmarking} & 30 &100  & 91,050 & \textbf{30} & 2-5 & 644,301 & $1920\times1080$ & Light Moving & Ground+Top \\
MDMT~\cite{liu2023robust} & 44 & 88  & 39,678 & - & 2 & 2,204,620 & $1920\times1080$ & Regular Moving & Aerial \\
MvMHAT(Self)~\cite{feng2025Unveiling}  & 26 & 98  & 90,900 & - & 3-4 & - & $2704\times1520$ & Light Moving & Ground \\
MMP-MvMHAT~\cite{feng2025Unveiling} & 2 & 10  & 12,000 & 15 & 4-6 & 85,650 & $640\times360$ & Static & Ground \\
DIVOTrack~\cite{hao2024divotrack} & 25 & 75  & 81,000 & \textbf{30} & 3 & 830,000 & $1920\times1080$ & Regular Moving & Ground \\
\hline
RealMvMoAT(ours) & \textbf{52} & \textbf{314} & \textbf{504,900} & \textbf{30} & \textbf{5--7} & \textbf{7,303,191} & $\textbf{3200}\times\textbf{1800}$ & \textbf{Irregular Moving} & \textbf{Ground+Aerial+Top} \\
\hline
\end{tabular*}
\vspace{-6mm}
\end{center}
\end{table*}

% \subsection{Data Annotation}
% We adopt a hybrid annotation pipeline. First, all camera views are temporally aligned through clipping and offset correction for fine-grained synchronization. Then, bounding box annotation varies across viewpoints: for top-down or near-top drone views, annotators label every five frames, followed by interpolation and manual refinement to address drift caused by camera motion or target movement. For oblique aerial and ground-level views, a YOLOv8X detector generates initial bounding boxes, which are manually verified and corrected.
% After that, identity annotation is carried out entirely manually. Annotators reference multiple views simultaneously using a custom annotation interface and are instructed to pay particular attention to challenging cases, including occlusion, partial visibility, similar clothing, and viewpoint-induced appearance changes.
% All annotations undergo multi-stage verification, including cross-view consistency checks, bounding box quality assessment under motion or blur, and identity validation in crowded scenes. This ensures high-quality and view-consistent annotations across the entire dataset.
% The full annotation workflow required over \textbf{1{,}500 man-hours}, including temporal alignment, detector-assisted labeling, identity assignment across views and time, and multi-stage verification.

\subsection{Data Annotation}
We adopt a hybrid annotation pipeline. First, all camera views are temporally aligned through clipping and offset correction for fine-grained synchronization. Bounding box annotation is then performed differently across viewpoints: for top-down or near-top drone views, annotators label every five frames, followed by interpolation and manual refinement to correct drift caused by camera motion or target movement; for oblique aerial and ground-level views, initial bounding boxes are generated by a YOLOv8X detector and then manually verified and corrected.
Identity annotation is performed entirely manually. Using a custom annotation interface, annotators inspect multiple views simultaneously and pay particular attention to challenging cases such as occlusion, partial visibility, similar clothing, and viewpoint-induced appearance changes.
All annotations undergo multi-stage verification, including cross-view consistency checks, bounding box quality assessment under motion or blur, and identity validation in crowded scenes, ensuring high-quality and view-consistent annotations across the dataset. The full workflow required over \textbf{1{,}500 man-hours}, covering temporal alignment, detector-assisted labeling, cross-view and temporal identity assignment, and multi-stage verification.

\subsection{Data Split, Evaluation Protocol, and Metrics}
To promote fair evaluation and emphasize generalization, we adopt a scene-level split: five scenes for training and five for testing. The two subsets share no overlap in location, camera configuration, or subject set, creating a significant domain gap that reflects real-world deployment, where unseen environments must be handled.

We provide two testing modes: GT-detection mode and Detector mode. In GT-detection mode, trackers use ground-truth bounding boxes; in Detector mode, users may apply any detector and evaluate against ground-truth annotations. All experiments conducted on the RealMvMoAT benchmark use the Detector mode.

Following standard MvMoAT evaluation practices~\cite{feng2025Unveiling}, we report three groups of metrics: cross-view association, temporal tracking, and overall performance. Cross-view association is evaluated by AF1 and Multi-view Multi-human Association Accuracy (MHAA), while temporal tracking is measured by IDF1, Multiple Object Tracking Accuracy (MOTA), and Higher Order Tracking Accuracy (HOTA). We further report two overall metrics: overall accuracy $\mathcal{A}=\mathrm{mean}(\mathrm{MHAA},\mathrm{MOTA})$ and overall F1 $\mathcal{F}=\mathrm{mean}(\mathrm{AF1},\mathrm{IDF1})$.
To assess spatial-temporal consistency over consecutive frames, we use Spatial-Temporal Matching Accuracy (STMA), denoted as $\mathcal{S}@N$. Specifically, AF1 is the F1 score of multi-view association precision and recall across all view pairs, and MHAA adopts a MOTA-style formulation that jointly penalizes missed subjects, false positives, and mismatches. For temporal tracking, IDF1 measures identity consistency via ID precision and recall, MOTA evaluates overall tracking accuracy, and HOTA balances detection, association, and localization performance. STMA computes the F1 score between predicted and ground-truth spatial-temporal matching matrices over consecutive frames, where $\mathcal{S}@5$, $\mathcal{S}@10$, and $\mathcal{S}@30$ evaluate short-, medium-, and long-term consistency.

\subsection{Comparison with Existing Benchmarks}
As summarized in Table~\ref{tab:comp_benchmark}, RealMvMoAT substantially advances existing MvMoAT datasets.
\textbf{Scale:} With 504K frames and over 7.3M bounding boxes, it is more than five times larger than any prior MvMoAT dataset.
\textbf{Camera mobility:} Unlike previous benchmarks that mainly use static, light, or regular moving cameras, RealMvMoAT includes \emph{Irregular Moving} cameras with random and unpredictable motion patterns, such as circular sweeps, hovering rotations, lateral panning, and zigzag flights. By comparison, \emph{Light Moving} involves gentle motions like slow panning or slight tilting, while \emph{Regular Moving} refers to steadier movements such as continuous panning or stable tracking.
\textbf{Viewpoint diversity:} It contains synchronized ground-level, oblique, top-down, and aerial views.
\textbf{Scene diversity:} It spans diverse open-space environments with varied densities, layouts, and lighting conditions.
These characteristics make RealMvMoAT a comprehensive and challenging benchmark for future multi-view association and tracking, especially in mobile or aerial--ground collaborative settings.

\subsection{Ethical Considerations and Privacy Protection}
The RealMvMoAT involves human participants. Informed consent was obtained from all participants, who were fully aware of the nature of the data collection, its intended use for pedestrian tracking, and the publication of identity-labeled annotations. To protect privacy, we blurred faces and other identifiable features that could reveal participants' identities in the real world. 
The data collection was approved by the relevant departments of our institution and complied with applicable laws and ethical guidelines for human-subject research. The dataset is publicly released under specific usage terms for research purposes only. No sensitive personal information was collected, and the data is intended solely for academic research on pedestrian tracking and multi-view data analysis.

\begin{figure*}[t] 
	\centering
	\includegraphics[width=0.875\linewidth,scale=0.875]{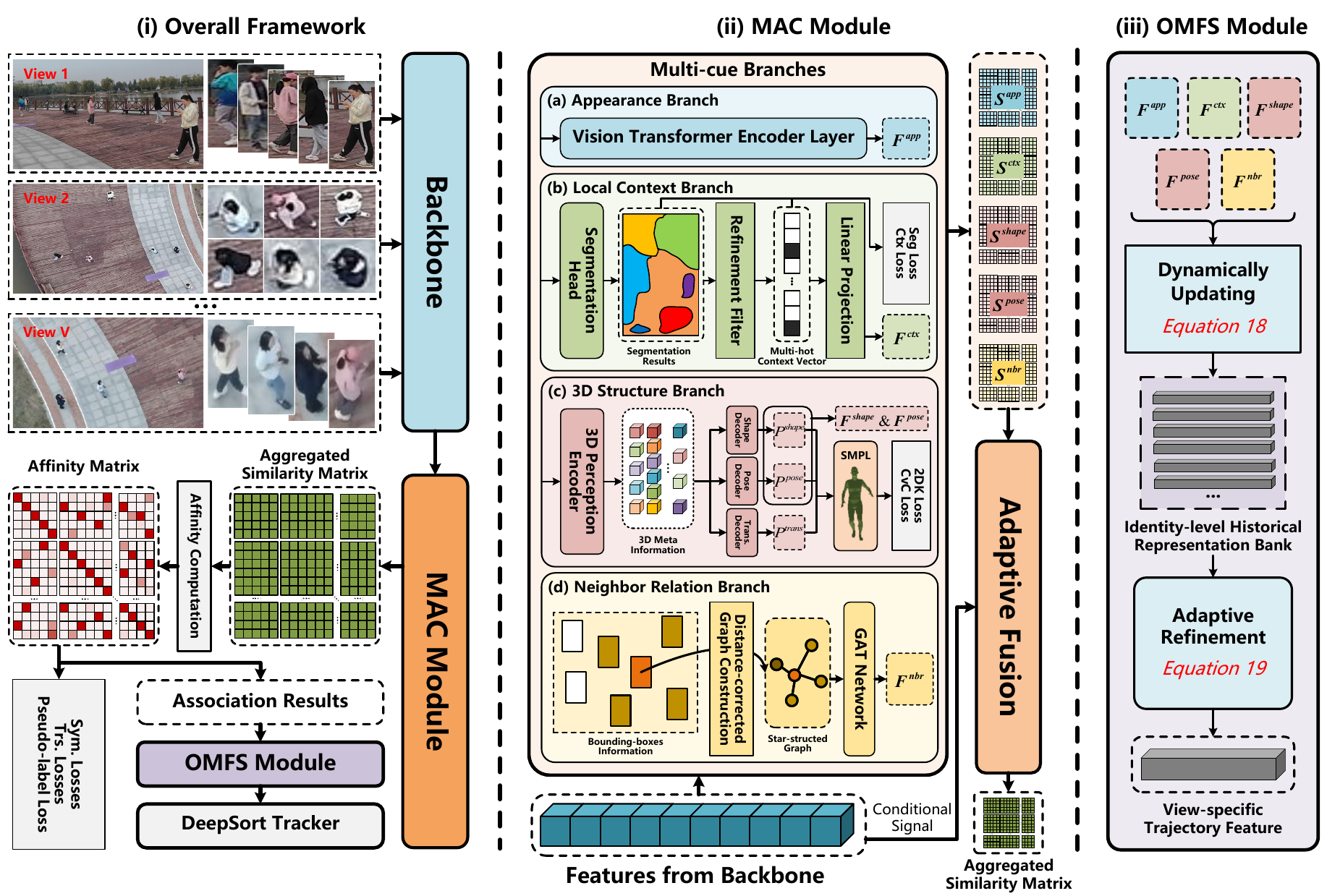}
    \vspace{-2mm}
	\caption{Overview of the proposed FUSION framework.
It consists of three parts: (i) the overall pipeline for MvMoAT, (ii) the MAC module that extracts and fuses multiple cues at the similarity level to produce the aggregated similarity matrix, and (iii) the OMFS module that maintains temporal identity memory and adaptively refines trajectory features for robust temporal tracking.
}
	\label{fig2}
\vspace{-4mm}
\end{figure*}

\section{Methodology}

\subsection{Formulation and Preliminary}
We formulate MvMoAT as solving two sub-problems: 
(1) associating the same pedestrian across views at each timestamp (cross-view association), and 
(2) maintaining identity continuity over time within each view (temporal tracking).

We adopt a simplified variant of the existing method MvMHAT*~\cite{feng2025Unveiling} as our baseline.
Given synchronized videos from $V$ views, let $\mathcal{B}_t=\{\mathcal{B}_{i,t}\}_{i=1}^V$ denote all detections at timestep $t$, where $\mathcal{B}_{i,t}=\{b^k_{i,t}\}_{k=1}^{N_{i,t}}$ is the set of detected boxes in view $i$. 
All boxes are encoded by a backbone $\Phi$, producing features for all targets at timestep $t$:
\begin{equation}
F_t = \Phi(\mathcal{B}_t).
\end{equation} 
Based on $F_t$, we compute a global assignment matrix $A_t \in \mathbb{R}^{(N_{1,t}+...+N_{V,t})^2}$ for cross-view association, followed by Hungarian matching to obtain ID assignments.
For temporal tracking, detections $B_{i,t}$ are associated with trajectories $\mathcal{T}_{t-1}$ from earlier frames. 
This process is expressed by a per-view matching matrix $\overline{A}^i_t \in \mathbb{R}^{M_{t-1}\times N_{i,t}}$, where $M_{t-1}$ is the number of active trajectories.

\subsection{Overall Framework}
We propose a \textbf{F}eature \textbf{U}nification framework for multi-view a\textbf{S}sociation and \textbf{I}dentificati\textbf{ON (FUSION)} to address MvMoAT under large viewpoint variations. 
As illustrated in Fig.~\ref{fig2}, FUSION consists of a feature-extraction backbone and two major modules: 
the Multi-cue Adaptive Combination (\textbf{MAC}) module for cross-view association, and 
the Online Multi-view Feature Synchronization (\textbf{OMFS}) module for temporal tracking.
MAC performs similarity-level adaptive fusion of multiple pedestrian cues, including appearance, local context, inherent 3D attributes, and neighbor relations, to produce unified cross-view affinity. 
The OMFS aggregates informative pedestrian features across historical frames and views to form robust trajectory representations, enabling accurate matching between trajectories and new detections and thereby enhancing tracking consistency.
Together, these components jointly optimize cross-view association and temporal tracking, alleviating the adverse effects caused by viewpoint variations.

\subsection{Multi-cue Adaptive Combination}

In MvMoAT, drastic viewpoint changes and frequent occlusions often cause severe cross-view appearance inconsistencies, making appearance-only association unreliable. To address this challenge, we integrate additional viewpoint-robust cues to improve the stability of similarity estimation. We therefore introduce the \textbf{Multi-cue Adaptive Combination (MAC)} module, which enhances the representation of each pedestrian by jointly exploiting multiple complementary cues.

After the shared backbone extracts initial features for each pedestrian instance, MAC processes these features through four parallel cue-specific branches together with an adaptive fusion mechanism. Each branch takes the hidden state from the penultimate transformer encoder layer and extracts a distinct cue that contributes to cross-view or temporal similarity estimation. The similarities predicted by all branches are then adaptively combined to produce a unified and more reliable similarity score. In the following, we describe the cues captured by each branch in detail.

\subsubsection{\textbf{Appearance Branch}}  
Appearance serves as primary cue for identity matching and provides the most general visual representation. We designate the last transformer encoder layer of the shared backbone as the appearance branch $\mathrm{MAC}_{app}$, which produces the appearance embedding for each pedestrian:
\begin{equation}
F^{\mathrm{app}} = L(F),
\end{equation}
where $F$ denotes the backbone feature and $L(\cdot)$ refers to the final transformer encoder layer. While appearance features are effective in many cases, their reliability degrades substantially under large viewpoint changes or heavy occlusions. This motivates the incorporation of additional complementary cues to provide more robust similarity estimation.

\subsubsection{\textbf{Local-context Branch}}
When appearance becomes unreliable due to severe occlusion or visually similar clothing, humans often rely on the surrounding environment to support identity inference. Prior studies~\cite{hayes2010neural,rice2013unaware} provide concrete evidence for this mechanism. 
% Hayes et al. demonstrated that the human hippocampus automatically encodes face--context associations, and recognition performance drops significantly when faces are presented with inconsistent backgrounds~\cite{hayes2010neural}. 
For example, Rice et al. showed that even when facial or body appearance is heavily degraded, participants can still recognize individuals above chance level by relying on scene and contextual information~\cite{rice2013unaware}. 
These findings indicate that environmental context can serve as a robust auxiliary cue when visual appearance is ambiguous. Motivated by these observations, the local-context branch extracts semantic information from the surrounding region of each pedestrian to enhance viewpoint-robust matching.

Concretely, we apply a lightweight segmentation head on penultimate hidden states to predict per-pixel semantic classes and then compress these predictions into a compact context embedding. We first reshape the token sequence $F$ into a spatial feature map. The segmentation head is defined as,
\begin{equation}
\label{eq:seg_pipeline}
M \;=\; \mathrm{Softmax}\!\Big(
    \mathrm{Head}\!\big(
        g_{\mathrm{gate}}\!\big(
            g_{\mathrm{ir}}\!\big(
                \phi_{\mathrm{p}}(\phi_{\mathrm{n}}(\mathrm{Reshape}(F)))
            \big)
        \big)
    \big)
\Big),
\end{equation}
where $\phi_{\mathrm{n}}$ denotes a per-channel layer normalization, and $\phi_{\mathrm{p}}$ is a lightweight projection module composed of a $1\times1$ convolution followed by BN, SiLU, and two $3\times3$ convolutions to inject local spatial context.
$g_{\mathrm{ir}}$ applies two inverted residual blocks~\cite{sandler2018mobilenetv2}, each performing channel expansion (expand factor $e=2$), depthwise convolution, and a squeeze–excitation step, yielding a refined feature representation.
$g_{\mathrm{gate}}$ is a squeeze–excitation–style channel gating module: it aggregates spatial information with global average pooling and generates a sigmoid gating vector to reweight feature channels.
Finally, $\mathrm{Head}$ is a $1\times1$ classifier over $C$ semantic classes, followed by a softmax applied along the class dimension.
The segmentation output is $M=[M_1,\dots,M_C]$, where each $M_c$ is a probability map, and each pixel $x$ is assigned the label $y(x)=\arg\max_{c} M_c(x)$.

We apply a refinement filter to suppress noise from imperfect segmentation and cross-view inconsistencies by binarizing each semantic category based on its area ratio:
\begin{equation}
\label{eq:refine_filter}
\begin{aligned}
r_c &= \frac{1}{|\Omega|} \sum_{x \in \Omega} \mathbf{1}\!\big[\, y(x) = c \,\big], \\
v_c &= \mathbf{1}\!\big[\, r_c > \tau_{\mathrm{area}} \,\big], \quad \tau_{\mathrm{area}} = 0.02
\end{aligned}
\end{equation}
where \(r_c\) is the normalized area ratio of semantic class \(c\), and 
\(v=[v_1,\dots,v_C]\in\{0,1\}^C\) is a \(C\)-dimensional multi-hot vector, with one entry per class. 
Here, \(\Omega\) denotes the pixel space of the feature map, \ie, the set of all positions in the feature map.

Finally, we map the refined context vector into a compact embedding:
\begin{equation}
\label{eq:ctx_embed}
F^{\mathrm{ctx}} \;=\; E_{\mathrm{ctx}}\, v, \qquad E_{\mathrm{ctx}} \in \mathbb{R}^{d_{\mathrm{ctx}}\times C}.
\end{equation}
Here, \(E_{\mathrm{ctx}}\) is a linear layer that projects the refined context vector \(v\) into a compact embedding \(F^{\mathrm{ctx}}\), where \(d_{\mathrm{ctx}}\) is the embedding dimension and \(C\) is the number of semantic classes. The resulting \(F^{\mathrm{ctx}}\) is then used for computing cross-view similarity scores.
This presence-based encoding provides stable, view-robust contextual evidence that helps disambiguate pedestrians under occlusion or appearance similarity while remaining insensitive to minor segmentation errors.

\subsubsection{\textbf{3D Structure Branch}}  
A person’s intrinsic 3D body information—such as skeletal pose and body shape—remains relatively stable across viewpoints at the same time, making it a natural auxiliary cue for cross-view association. Neuroscience and perception studies~\cite{foster2021separated,johansson1973visual} support this property. 
% Foster et al. showed that body-selective regions in the ventral visual stream encode viewpoint-invariant body identity representations~\cite{foster2021separated}.
For example, Johansson’s classical motion experiments demonstrated that humans can reliably recognize individuals from sparse joint movements alone~\cite{johansson1973visual}. 
These findings indicate that 3D structural cues provide perceptually stable signatures that persist under viewpoint changes, motivating their integration into our multi-cue design.

To leverage this cue, the 3D structure branch predicts coarse SMPL-based pose and shape parameters from the shared backbone features. Given the extracted feature set $F_i$ for the $N_i$ detected pedestrians under view~$i$, an encoder $\sigma(\cdot)$, implemented as a lightweight two-layer MLP, maps them into intermediate 3D meta information:
\begin{equation}
f_i = \sigma(F_i).
\end{equation}
Three lightweight decoders, each a two-layer MLP, then estimate pose, shape, and weak-perspective translation parameters (where translation is used only internally for SMPL reprojection and is not preserved as part of the final representation):
\begin{equation}
P^{\text{pose/shape/trans}}_i = DEC_{\text{pose/shape/trans}}(f_i),
\end{equation}
where $P^{\text{pose}}_i \in \mathbb{R}^{N_i \times 24 \times 3}$, 
$P^{\text{shape}}_i \in \mathbb{R}^{N_i \times 10}$, 
and $P^{\text{trans}}_i \in \mathbb{R}^{N_i \times 3}$.

During training, inspired by the self-supervised consistency strategy of RSC-Net~\cite{xu20213d}, we impose multi-view consistency constraints that encourage predicted pose and shape to remain aligned across synchronized views, and additionally require their 2D projections to agree with annotated keypoints when available. Detailed loss formulations are provided in a later subsection. At inference time, only the predicted pose and shape are retained as auxiliary cues for adaptive fusion:
\begin{equation}
F^{\text{pose}}_i = P^{\text{pose}}_i, 
\qquad
F^{\text{shape}}_i = P^{\text{shape}}_i.
\end{equation}

\subsubsection{\textbf{Neighbor Relation Branch}}
In synchronized multi-view scenes, the relative arrangement of a pedestrian and the people around it remains largely consistent across viewpoints, making local adjacency a stable relational cue for cross-view association. Prior studies support this intuition: Yan et al. demonstrated that explicitly modeling inter-person spatial dependencies through graph-based attention substantially improves identity reasoning in multi-person environments~\cite{yan2020learning}. Such findings indicate that group-level spatial structure provides transferable identity cues, especially when individual appearance is ambiguous. Motivated by this, the neighbor relation branch captures the local geometric configuration around each pedestrian as an auxiliary cue.

Concretely, we construct a star-structured pedestrian graph $\mathcal{G}=(\mathcal{V},\mathcal{E})$ in each frame, where each pedestrian is the center node connected to its nearest $N\!=\!4$ neighbors. The connection strength (edge between nodes) is determined by a height-adjusted center distance, approximating geometric proximity in perspective-view imagery. Let $c_a, c_b$ denote bounding-box centers and $h_a, h_b$ their heights. We define the base distance $d^{(0)}_{ab}=\|c_a - c_b\|_2$ and the normalized height discrepancy
\begin{equation}
z_{ab} \;=\; \frac{|h_a - h_b|}{h_{\mathrm{ref}}},
\end{equation}
where $h_{\mathrm{ref}}$ is a per-view normalization constant computed as the median of all pedestrian bounding-box heights in that view.
A threshold-gated correction is applied only when the discrepancy exceeds $\tau_{\mathrm{height}}$:
\begin{equation}
\label{eq:height_scaled_dist}
d_{ab} \;=\; d^{(0)}_{ab}\,\Big(1+\lambda\,\big[z_{ab}-\tau_{\mathrm{height}}\big]_+\Big),
\qquad [x]_+ = \max(0,x),
\end{equation}
i.e.,
\begin{equation}
d_{ab} \;=\;
\begin{cases}
d^{(0)}_{ab}, & z_{ab} \le \tau_{\mathrm{height}},\\[3pt]
d^{(0)}_{ab}\!\left(1+\lambda\,(z_{ab}-\tau_{\mathrm{height}})\right), & z_{ab} > \tau_{\mathrm{height}},
\end{cases}
\end{equation}
where $\lambda$ controls the influence of height disparity. This formulation preserves proximity for minor height variations while increasing distances only when the discrepancy indicates meaningful depth differences, improving geometric reliability in ground-view scenes.

We bias message passing towards spatially reliable neighbors by converting the adjusted distance into a normalized proximity weight:
\begin{equation}
w_{ab} = \exp(-\,\gamma\, d_{ab}),
\end{equation}
where $\gamma$ controls the decay rate, these weights serve as soft geometric priors that encourage stronger interactions with closer and more reliable neighbors.

The resulting graph is processed using a Graph Attention Network v2 (GATv2)~\cite{brody2021attentive}, which learns attention coefficients for each edge while incorporating the proximity weights. Node features are initialized with the backbone embeddings of each pedestrian, and edge features are initialized using the proximity weights $w_{ab}$, which serve as edge-level attention biases. For each center node, GATv2 performs attention-based aggregation over its $N$ neighbors:
\begin{equation}
F^{\mathrm{nbr}} = \mathrm{GATv2}(\mathcal{G}),
\end{equation}
yielding neighbor-aware embeddings that encode both geometric proximity and learned relational importance. These features complement other branches by providing stable structural cues that enhance identity matching in crowded, occluded, or visually ambiguous scenarios.

\subsubsection{\textbf{Adaptive Fusion}}
Given the heterogeneous features extracted from the four branches, we perform cue fusion at the similarity level. We adopt a time-synchronized and cross-view shared fusion strategy: for each timestamp $t$, all pedestrians across all views share the same set of fusion weights $\alpha_t$. This design enforces a unified cue-combination policy within each frame, ensuring that fusion behavior remains consistent across different viewpoints observing the same moment.

Formally, let $F^{m}_{i,k,t}$ denote the feature of branch 
$m \in \{\mathrm{app}, \mathrm{ctx}, \mathrm{pose}, \mathrm{shape}, \mathrm{nbr}\}$ 
for the $k$-th pedestrian under view $i$ at time $t$. Let 
$F_{i,k,t}$ be the corresponding backbone feature from the penultimate transformer encoder layer, which serves as the shared input to all branch-specific heads. To capture the global scene condition at time $t$, we compute a frame-level descriptor by averaging the backbone features across all views and all detected pedestrians:
\begin{equation}
\bar{F}_{t} \;=\;
\frac{1}{\sum_{i=1}^{V} N_{i,t}}
\sum_{i=1}^{V} \sum_{k=1}^{N_{i,t}} F_{i,k,t}.
\end{equation}
A lightweight generator $\psi(\cdot)$ then produces non-negative normalized fusion weights:
\begin{equation}
\alpha_{t} \;=\; \mathrm{Softmax}\!\big(\psi(\bar{F}_{t})\big),
\qquad
\sum_{m}\alpha_{t,m}=1.
\end{equation}

For cross-view association, we compute a similarity matrix per cue. Given two pedestrians $u$ and $v$ (possibly from different views) at time $t$, the cue-specific cosine similarity is
\begin{equation}
S^{m}_{t}(u,v)
\;=\;
\frac{\langle F^{m}_{u,t},\,F^{m}_{v,t}\rangle}
{\|F^{m}_{u,t}\|_2\,\|F^{m}_{v,t}\|_2}.
\end{equation}
To mitigate scale mismatches across cues, each $S^{m}_{t}$ is normalized with a branch-specific operator, denoted by
$\tilde{S}^{m}_{t}=\mathrm{Norm}_{m}(S^{m}_{t})$.
The final fused similarity is obtained via a weighted sum:
\begin{equation}
S^{*}_{t}
\;=\;
\sum_{m \in 
\{\mathrm{app},\mathrm{ctx},\mathrm{pose},\mathrm{shape},\mathrm{nbr}\}}
\alpha_{t,m}\,\tilde{S}^{m}_{t}.
\end{equation}

This adaptive fusion mechanism preserves the independence of individual cues while enforcing a unified frame-level weighting scheme shared across all views. This design improves robustness under viewpoint shifts, occlusions, and appearance ambiguities by dynamically adjusting cue contributions based on scene conditions at each timestamp.

\subsection{Online Multi-view Feature Synchronization}
After the MAC module extracts diverse cue features, we further match them with trajectory features to enable robust tracking over time. In RealMvMoAT, we reflect an important real-world challenge: the dynamic motion of the capturing platform causes pedestrian viewpoints to change continuously, leading to substantial inter-frame divergence that hinders matching between new detections and historical trajectories. 
To address this challenge, we leverage both historical and cross-camera information to construct robust pedestrian trajectory representations, where features from different cues and viewpoints are temporally aggregated and refined into a unified identity memory for accurate matching with new detections.
The motivation of this idea is intuitive: by integrating representations of the same pedestrian observed under diverse viewpoints across time and cameras, the trajectory features could become more expressive and robust to the viewpoint present in incoming frames~\cite{quiroga2005invariant}.

To accomplish this goal, we propose the Online Multi-view Feature Synchronization (OMFS) module, which aggregates historical representations of each pedestrian identity across all views to enhance trajectory features. 
% In this module, cue embeddings from the four MAC branches are used independently for tracking. A temporal identity memory $\mathcal{M}$ preserves cross-time and cross-view features for each identity. For the $i$-th pedestrian at timestep $t$, $\mathcal{M}^b_{i,t}$ stores its $b$-th cue information from historical frames under all views, where $b \in \{\text{app}, \text{ctx}, \text{pose}, \text{shape}, \text{nbr}\}$. $\mathcal{M}^b_{i,t}$ is initialized with zeros and updated using a momentum strategy:
In this module, cue embeddings from the four MAC branches are used independently for tracking. A temporal identity memory \(\mathcal{M}\) preserves cross-time and cross-view features for each identity. For the \(i\)-th pedestrian at timestep \(t\), \(\mathcal{M}^b_{i,t}\) stores its \(b\)-th cue information from historical frames under all views, where \(b \in \{\text{app}, \text{ctx}, \text{pose}, \text{shape}, \text{nbr}\}\). \(\mathcal{M}^b_{i,t}\) is initialized as an empty memory and updated by appending the current cue embedding to the identity-level historical representation memory bank:

\begin{equation}
\begin{aligned}
\mathcal{M}^{b}_{i,t} \;=\;
\begin{cases}
0, & t=0,\\[3pt]
[\mathcal{M}^b_{i,t-1}, \mathrm{Avg}\!\big(F^{b}_{i,t}\big)] \in \mathbb{R}^{L \times d}, & t > 0,
\end{cases} &\\
\end{aligned}
\end{equation}
where \(F^b_{i,t}\) is the embedding features of the \(b\)-th cue belonging to the \(i\)-th pedestrian at timestep \(t\), and \(L=100\) is the maximum length of the memory bank. Once the memory bank reaches the maximum length \(L\), the oldest entry is discarded. This design preserves historical information while ensuring that the memory bank does not grow too large over time.

Thereafter, we use a cross-attention operation to refine the trajectory features with the temporal identity memory \(\mathcal{M}\). Take the \(b\)-th cue as an example. At timestep \(t\), the trajectory feature of the \(i\)-th pedestrian \(T^{b,v}_{i,t}\) under camera-view \(v\) is used as the query, and its corresponding memory feature at timestep \(t-1\), i.e., \(\mathcal{M}^{b}_{i,t-1}\), is used as the key and value. Formally, this process is written as:
\begin{equation}
\tilde{T}^{b,v}_{i,t} = \mathrm{Softmax}\left(\frac{T^{b,v}_{i,t} \cdot (\mathcal{M}^{b}_{i,t-1})^\top}{\sqrt{d}}\right) \mathcal{M}^{b}_{i,t-1}
\end{equation}
where \(d\) is the feature dimension of the \(b\)-th cue; \(\tilde{T}^{b,v}_{i,t}\) is the refined trajectory feature of the \(b\)-th cue. This design allows the model to refine the trajectory features of each pedestrian instance using their historical and cross-view information.

During temporal tracking, we use the refined trajectory embeddings of each cue to compute similarity with corresponding cue embeddings of targets in incoming frames, resulting in 5 similarity matrices. These matrices are averaged to form the final similarity matrix between trajectory features and newly detected targets. We then apply DeepSORT~\cite{wojke2017simple} to compute the optimal assignment between historical trajectories and newly detected targets based on this similarity matrix.

\subsection{Optimization}

For training and optimization, we follow the same self-supervised learning paradigm as the MvMHAT*~\cite{feng2025Unveiling}. The basic objective is
\begin{equation}
\mathcal{L}_\text{basic} = \mathcal{L}^A_\text{Sym} + \mathcal{L}^A_\text{Trs} + \mathcal{L}^M_\text{Pse} + \mathcal{L}^M_\text{Sym} + \mathcal{L}^M_\text{Trs},
\end{equation}
where $\mathcal{L}^A_\text{Sym}$ and $\mathcal{L}^A_\text{Trs}$ denote the symmetric- and transitive-consistency losses (Sym loss and Trs loss) for appearance feature learning, and $\mathcal{L}^M_\text{Sym}$ and $\mathcal{L}^M_\text{Trs}$ are the corresponding terms for cross-view assignment learning. $\mathcal{L}^M_\text{Pse}$ is a pseudo-label loss supervising the cross-view assignment with labels obtained by the Hungarian algorithm from the fused similarity $S^{*}_t$. 
The detailed formulations of these losses can be found in ~\cite{feng2025Unveiling}.

In addition to $\mathcal{L}_\text{basic}$, several auxiliary losses are introduced to facilitate the training of individual branches.

\subsubsection{\textbf{Local-context branch}}
For the local-context branch, we adopt two auxiliary losses to enhance background modeling and cross-view consistency. 
We employ a segmentation loss on the per-pixel predictions, using teacher-provided pseudo labels as supervision, guiding the model to learn meaningful contextual segmentation around each target. Let $\hat{Y}\in\{1,\dots,C\}^{H\times W}$ denote the pseudo labels and $M$ the predicted probability maps (Eq.~\ref{eq:seg_pipeline}). The segmentation loss (Seg Loss) is
\begin{equation}
\mathcal{L}_\text{seg} \;=\; - \frac{1}{|\Omega_{\text{img}}|}\sum_{x\in\Omega_{\text{img}}}\log M_{\hat{Y}(x)}(x).
\end{equation}
To further reduce view-specific noise, we regularize the refined context presence vectors across views, encouraging the model to predict consistent semantic context for images of the same pedestrian simultaneously captured under diverse camera views. Let $v_{i,t}$ and $v_{j,t}$ be the binary vectors (Eq.~\ref{eq:refine_filter}) from views $i$ and $j$ at time $t$. The context-consistency loss (Ctx Loss) via consistency regularization is
\begin{equation}
\mathcal{L}_\text{ctx} \;=\; \frac{2}{V(V-1)} \sum_{i=1}^{V}\sum_{j=i+1}^{V} \| v_{i,t} - v_{j,t} \|_1.
\end{equation}

\subsubsection{\textbf{3D structure branch}}
The 3D structure branch is supervised by both keypoint and multi-view consistency objectives. 
A 2D keypoint loss $\mathcal{L}_\text{2dkey}$ (2DK Loss) constrains the reprojected 3D joints to align with annotated 2D keypoints when available, ensuring accurate estimation of human pose and shape parameters:
\begin{equation}
\mathcal{L}_\text{2dkey} = \frac{1}{V} \sum_{i=1}^{V} \| J^{2d}_i - \hat{J}^{2d}_i \|^2 ,
\end{equation}
where $J^{2d}_i$ is the projection of predicted 3D joints under view-$i$,and the 2D labels are obtained by ViTPose~\cite{xu2022vitpose}.

To encourage consistency across views, we introduce two regularizers. First, a parameter consistency loss $\mathcal{L}_{\text{consis-smpl}}$ forces the model to predict consistent pose and parameter for images of the same pedestrian simultaneously captured under diverse camera views:
\begin{equation}
\mathcal{L}_{\text{consis-smpl}} = \tfrac{2}{V(V-1)} \sum_{i<j} \big( \| P^\text{pose}_{i} - P^\text{pose}_{j} \|^2 + \| P^\text{shape}_{i} - P^\text{shape}_{j} \|^2 \big).
\end{equation}
Second, a 3D keypoints consistency loss $\mathcal{L}_{\text{consis-3dkey}}$ ensures the joints of 3D models reconstructed from images of the same pedestrian are correctly aligned after rigid Procrustes alignment $\mathcal{A}(\cdot)$:
\begin{equation}
\mathcal{L}_{\text{consis-3dkey}} = \tfrac{2}{V(V-1)} \sum_{i<j} \| \mathcal{A}(J^{3d}_{i}) - \mathcal{A}(J^{3d}_{j}) \|^2 .
\end{equation}

The total cross-view consistency loss (CvC Loss) is denoted as:
\begin{equation}
\mathcal{L}_{\text{consis}} = \mathcal{L}_{\text{consis-smpl}} + \mathcal{L}_{\text{consis-3dkey}} .
\end{equation}

\subsubsection{\textbf{Neighbor relation branch}}
No additional supervision is introduced; the neighbor features are implicitly optimized through the overall training objective.

\subsubsection{\textbf{Overall objective}}
The final training loss combines all terms:
\begin{equation}
\mathcal{L} \;=\; \mathcal{L}_\text{basic}
\;+\; \beta_1 \mathcal{L}_\text{seg}
\;+\; \beta_2 \mathcal{L}_\text{ctx}
\;+\; \beta_3 \mathcal{L}_\text{2dkey}
\;+\; \beta_4 \mathcal{L}_\text{consis},
\label{eq:overall_loss}
\end{equation}
where $\beta_1,\beta_2,\beta_3,\beta_4$ are trade-off weights.

\section{Experiments}

\subsection{Benchmarks and Metrics}

\subsubsection{\textbf{Benchmarks}} 
We evaluate our approach on both the proposed RealMvMoAT benchmark and several public multi-view datasets. RealMvMoAT provides large-scale real-world data with diverse viewpoints and mobile platforms, offering a realistic evaluation setting. In addition, we conduct experiments on six widely used benchmarks, including MvMHAT~\cite{feng2025Unveiling}, CvMHAT-R~\cite{han2024benchmarking}, MMP-MvMHAT~\cite{han2021mmptrack}, MvMHAT (Self)~\cite{feng2025Unveiling}, EPFL~\cite{fleuret2008multicamera}, and Campus~\cite{xu2016multi}, covering a variety of indoor and outdoor environments with multiple synchronized cameras. WildTrack~\cite{chavdarova2018wildtrack} is omitted due to annotation inaccuracies reported in prior work~\cite{hao2024divotrack,teepe2024lifting,alturki2025enhanced}. Following standard protocols, ground-truth detections are used for training, while inference relies on YOLOv8~\cite{Glenn2023yolov8} and Detectron2~\cite{wu2019detectron2} for RealMvMoAT and MvMHAT, and ground-truth detections for CvMHAT-R and MMP-MvMHAT. Additional supervision is provided by 2D keypoints and segmentation masks generated using ViTPose~\cite{xu2022vitpose,xu2023vitpose++} and MaskCLIP+~\cite{zhou2022extract}.

\subsubsection{\textbf{Metrics}}
Performance is evaluated using the standard MvMoAT metrics~\cite{feng2025Unveiling}, which cover cross-view association (AF1, MHAA), temporal identity tracking (IDF1, MOTA, HOTA~\cite{luiten2021hota}), and overall indicators ($\mathcal{A}$, $\mathcal{F}$, and $\mathcal{S}@N$).

\subsection{Implementation Details}

We use the simplified "ViT+FC" variant of the self-supervised multi-view framework MvMHAT*~\cite{feng2025Unveiling} as our baseline, with ViT-S/16~\cite{dosovitskiy2020image} as the backbone and a lightweight four-layer fully connected network for the assignment module. Temporal tracking follows the DeepSORT pipeline~\cite{wojke2017simple}, using cascade matching and an IOU/ReID hybrid association strategy. To improve the stability of the 3D structure branch, we initialize its encoder and decoders with pretrained weights from the human mesh recovery model RSC-Net~\cite{xu20213d}, following common practice in mesh reconstruction~\cite{kanazawa2018end}.

For training, all input bounding boxes are resized to $224\times224$. The model is trained for 15 epochs with an initial learning rate of $10^{-5}$. The loss trade-off parameters in Eq.~\eqref{eq:overall_loss} are set to $\beta_1 = 0.5$, $\beta_2 = 0.25$, $\beta_3 = 0.5$, and $\beta_4 = 0.25$. All experiments are implemented in PyTorch and conducted on a workstation equipped with two NVIDIA RTX A6000 GPUs.

\begin{table*}[t]
\centering
\caption{The experiment results of our FUSION and other state-of-the-art methods on RealMvMoAT. }
\vspace{-3mm}
\label{tab:real}

\small
\renewcommand\arraystretch{0.90}
\setlength{\tabcolsep}{9.8pt}
\begin{tabular*}{\textwidth}{ l | c c | c c c | c c c c c}
    \hline
        \multirow{2}*{Variations} & \multicolumn{2}{c|}{Cross-view } & \multicolumn{3}{c|}{Temporal} & \multicolumn{5}{c}{Overall} \\ \cline{2-11}
         & AF1 & MHAA & IDF1 & MOTA & HOTA & $\mathcal{A}$ & $\mathcal{F}$ & $\mathcal{S}$@5 & $\mathcal{S}$@10 & $\mathcal{S}$@30\\ \hline\hline
   
        Tracktor++~\cite{bergmann2019tracking} & 18.8  & 23.5  & 37.8  & 38.9  & 22.5  & 31.2  & 28.3  & 32.6  & 31.7  & 30.5  \\ 
        CenterTrack ~\cite{zhou2020tracking} & 10.3  & 19.8  & 28.2  & 35.5  & 19.2  & 27.7  & 19.2  & 28.3  & 27.3  & 25.5  \\ 
        TraDeS ~\cite{wu2021track} & 17.0  & 22.4  & 36.0  & 40.1  & 22.2  & 31.3  & 26.5  & 34.1  & 33.3  & 31.9  \\ 
        TrackFormer ~\cite{meinhardt2022trackformer} & 22.1  & 23.2  & 36.6  & 38.6  & 22.7  & 30.9  & 29.3  & 35.3  & 34.5  & 33.2  \\ 
        SSMOT ~\cite{bastani2021self}+Ref ~\cite{gan2021self} & 30.5  & 27.0  & 40.5  & 42.1  & 25.7  & 34.5  & 35.5  & 39.0  & 38.0  & 36.7  \\  
        MvMHAT*~\cite{feng2025Unveiling} & 55.1  & 41.0  & 60.8  & 59.2  & 41.7  & 50.1  & 58.0  & 52.5  & 51.1  & 49.4  \\ \hline
        baseline & 51.7 & 38.1 & 58.7 & 59.3 & 40.8 & 48.7 & 55.2 & 50.0 & 49.1 & 48.0  \\ 
        % Ours & 60.9  & 45.7  & 69.7  & 68.3  & 47.1  & 57.0  & 65.3  & 61.3  & 59.8  & 58.5  \\ \hline
        Ours & \textbf{65.7}  & \textbf{49.8}  & \textbf{74.4}  & \textbf{72.9}  & \textbf{50.3}  & \textbf{61.4}  & \textbf{70.5}  & \textbf{62.5}  & \textbf{61.1}  & \textbf{60.0}  \\ \hline
\end{tabular*}

\vspace{-3mm}
\end{table*}

\subsection{Comparison with State-of-the-art Methods}

We compare our FUSION framework with several state-of-the-art approaches on multiple benchmarks, showing that FUSION consistently outperforms others in both cross-view association and temporal tracking metrics. In addition to the results on RealMvMoAT, more results on other public benchmarks are provided in the \textbf{Supplementary Materials}.

\subsubsection{\textbf{Comparison on Our RealMvMoAT Benchmark.}} 
As shown in Table~\ref{tab:real}, FUSION surpasses the top-performing method MvMHAT*~\cite{feng2025Unveiling} by a large margin, \ie, 10.9\% on average. MvMHAT* shares a similar structure with our baseline but replaces the identity assignment module with a tailored module to achieve superior performance. 
These results highlight the effectiveness of our MAC, which fuses appearance, local context, 3D structure, and neighbor cues at the similarity level, together with the OMFS, which aggregates information across views and frames for stable identity association. 
As a result, FUSION produces fewer identity switches and more stable trajectories in complex real-world multi-view settings.

\subsubsection{\textbf{Comparison on Public MvMoAT Benchmarks}} 
We report additional results on three standard MvMoAT benchmarks, MvMHAT, CvMHAT-R, and MMP-MvMHAT. As shown in Tables \uppercase\expandafter{\romannumeral7} -- \uppercase\expandafter{\romannumeral9} (in Supplementary Materials), our method consistently outperforms existing approaches across all major metrics, including AF1, MHAA, IDF1, MOTA, and HOTA. The improvements are in line with the trends observed on RealMvMoAT: by integrating multi-cue similarity estimation with cross-view feature synchronization, our framework yields more reliable identity association and more stable temporal trajectories than prior methods. The consistency of these gains across datasets with different camera arrangements and viewpoint variations further demonstrates the robustness and generalization of our design.
% From the results, we observe a salient finding that the performance gain of our FUSION model over compared methods is significantly larger on RealMvMoAT than on other datasets. This could be attributed to the higher diversity of viewpoints and stronger appearance variations in our dataset caused by moving cameras, which further demonstrates FUSION’s robustness under dynamic aerial–ground scenarios.

To ensure a fair comparison with recent approaches, we conduct experiments on three public benchmarks: MvMHAT (Self), EPFL, and Campus, as shown in Table \uppercase\expandafter{\romannumeral10} (in Supplementary Materials). FUSION outperforms the top competitor, CrossMOT~\cite{hao2024divotrack}, achieving higher AF1 and MHAA scores (following CrossMOT’s metric setting), except for MHAA@EPFL. These results highlight that our approach excels on large-scale benchmarks and generalizes well to smaller multi-view tracking datasets with static camera setups.

\subsection{Ablation Studies}
In this subsection, we perform a series of ablation studies to validate the effectiveness of the proposed FUSION framework. We first investigate the contribution of each component, including the MAC module, the OMFS module, and different cue branches, on the RealMvMoAT and MvMHAT benchmarks. Then, we analyze the impact of the trade-off parameters in the overall loss function. Next, we examine the necessity of the key designs in the neighbor relation branch, including the height-based distance calibration and the GATv2-based relational modeling. Moreover, we compare the computational complexity and latency of our method with the baseline. Finally, we present qualitative visualization results to further illustrate the advantages of our proposed multi-cue design in challenging multi-view association scenarios.

\begin{table*}[t]
\centering
\caption{Ablation study of proposed components and detailed setups on RealMvMoAT sub-set.}
\vspace{-3mm}
\label{tab:abl}

\small
\renewcommand\arraystretch{0.90}
\setlength{\tabcolsep}{8.15pt}
\begin{tabular*}{\textwidth}{ l | c c | c c c | c c c c c}
%            ^^^^^^^^^^^^  ^^^^^^^^^^^^^^^^^^^^
%            总宽度 = \textwidth   自动拉伸列间距填满
    \hline
    \multirow{2}*{Variations} & \multicolumn{2}{c|}{Cross-view } &
    \multicolumn{3}{c|}{Temporal} & \multicolumn{5}{c}{Overall} \\ \cline{2-11}
     & AF1 & MHAA & IDF1 & MOTA & HOTA & $\mathcal{A}$ & $\mathcal{F}$ &
       $\mathcal{S}$@5 & $\mathcal{S}$@10 & $\mathcal{S}$@30 \\ \hline\hline
        
        Baseline & 45.9  & 32.4  & 48.4  & 49.0  & 27.9  & 40.7  & 47.2  & 43.6  & 42.1  & 40.8  \\ \hline
        + 3D Structure Branch w/o Pose & 49.0  & 35.2  & 49.0  & 49.8  & 28.5  & 42.5  & 49.0  & 45.1  & 43.9  & 42.9   \\
        + 3D Structure Branch w/o Shape & 50.5  & 36.7  & 49.6  & 50.8  & 29.6  & 43.8  & 50.1  & 46.5  & 45.6  & 44.2   \\
        + 3D Structure Branch & 52.9  & 38.5  & 50.7  & 51.6  & 30.7  & 45.1  & 51.8  & 47.8  & 46.3  & 45.2   \\
        + Local-context Branch & 49.8  & 35.8  & 49.6  & 50.9  & 29.9  & 43.4  & 49.7  & 46.3  & 44.8  & 43.0   \\
        + Neighbor Relation Branch & 49.5  & 35.7  & 50.1  & 50.8  & 29.0  & 43.3  & 49.8  & 45.9  & 44.7  & 43.2 \\
        + MAC w/o Adaptive Fusion  & 54.3  & 39.8  & 51.2  & 52.2  & 31.3  & 46.0  & 52.8  & 49.3  & 47.8  & 46.7  \\ 
        + MAC  & 55.2  & 40.4  & 51.5  & 52.5  & 31.6  & 46.5  & 53.4  & 50.6  & 48.5  & 47.3  \\ 
        \hline
        + OMFS & 46.0  & 32.7  & 53.6  & 54.0  & 33.2  & 43.4  & 49.8  & 47.7  & 46.2  & 45.4   \\
        % + MAC + OMFS w/o Multi-cue Information & 55.5  & 40.9  & 56.7  & 56.8  & 35.3  & 49.4  & 56.1  & 51.8  & 50.5  & 49.3   \\
        + MAC + OMFS (Ours) & 55.5  & 40.7  & 62.0  & 60.9  & 39.8  & 50.8  & 58.8  & 52.7  & 51.3  & 49.6   \\
        \hline
        MvMHAT* & 48.6  & 34.7  & 49.5  & 50.6  & 28.4  & 42.7  & 49.1  & 45.9  & 44.0  & 42.6   \\ \hline
\end{tabular*}

\vspace{-3mm}
\end{table*}

\subsubsection{\textbf{Effectiveness of Proposed Components}}
% \noindent\textbf{Effectiveness of Proposed Components.}
To evaluate the effectiveness of our proposed MAC module and OMFS module, we add several components to the baseline gradually and evaluate the performances. In addition to the results on RealMvMoAT, more results on public MvMHAT benchmark are provided in the \textbf{Supplementary Materials}.

\textbf{Results on RealMvMoAT.}
Table~\ref{tab:abl} presents ablation results on RealMvMoAT for each component of our framework. Experiments are run on a RealMvMoAT sub-set for efficiency while preserving representative scene diversity.    
We begin with a baseline, a simplified variant of the existing method MvMHAT*, using only appearance features. Adding individual branches demonstrates their distinct contributions: the 3D structure branch yields the largest gain due to its view-invariant geometric cues (``+ 3D Structure Branch''), with pose information (``+ 3D Structure Branch w/o Shape'') contributing more than shape (``+ 3D Structure Branch w/o Pose''). The local-context (``+ Local-context Branch'') and neighbor-relation (``+ Neighbor Relation Branch'') branches provide additional improvements by enhancing robustness under appearance confusions. In addition, we observe that simply averaging the similarities across different cues (``+ MAC w/o Adaptive Fusion'') provides a modest performance improvement, whereas our adaptive fusion achieves greater gains, highlighting the effectiveness of dynamically balancing heterogeneous cues.
For temporal tracking, OMFS (``+ OMFS'') substantially enhances IDF1, MOTA, and HOTA by maintaining identity-level memory. Its combination with MAC leads to further improvement (``MAC + OMFS''), indicating that spatial multi-cue discrimination and temporal identity consistency complement each other effectively.

\textbf{Results on MvMHAT.}
As shown in Table~\uppercase\expandafter{\romannumeral11} (in Supplementary Materials), we further conduct the same ablations on the public MvMHAT benchmark and observe consistent trends that match those on RealMvMoAT:
3D cues improve view-invariant matching, local-context and neighbor cues enhance robustness under occlusion and crowding, adaptive fusion offers consistent gains over uniform fusion, and OMFS stabilizes temporal identity association. The full MAC + OMFS configuration achieves the highest overall performance, confirming that all proposed components generalize reliably across multi-view benchmarks.
The results on MvMHAT confirm the generality and reliability of each component.

\begin{table*}[t]
\centering
\caption{Ablation study on the trade-off parameters $\beta_1$--$\beta_4$ of the overall loss function on the RealMvMoAT sub-set.}
\vspace{-3mm}
\label{tab:tradeoff}

\small
\renewcommand\arraystretch{0.90}
\setlength{\tabcolsep}{6.85pt}
\begin{tabular*}{\textwidth}{ l | c c c c | c c | c c c | c c c c c }
    \hline
        \multirow{2}*{Variations} 
        & \multirow{2}*{$\beta_1$} & \multirow{2}*{$\beta_2$} & \multirow{2}*{$\beta_3$} & \multirow{2}*{$\beta_4$} 
        & \multicolumn{2}{c|}{Cross-view} 
        & \multicolumn{3}{c|}{Temporal} 
        & \multicolumn{5}{c}{Overall} \\ \cline{6-15}
        & & & & 
        & AF1 & MHAA 
        & IDF1 & MOTA & HOTA 
        & $\mathcal{A}$ & $\mathcal{F}$ & $\mathcal{S}$@5 & $\mathcal{S}$@10 & $\mathcal{S}$@30 \\ \hline\hline
1 (Ours) & 0.5 & 0.25 & 0.5 & 0.25 & 55.6 & 40.7 & 62.0 & 60.9 & 39.8 & 50.8 & 58.8 & 52.7 & 51.3 & 49.6 \\
2 & 1.0 & 0.25 & 0.5 & 0.25 & 55.6 & 40.0 & 62.2 & 60.7 & 39.9 & 50.4 & 58.9 & 53.2 & 51.3 & 49.4 \\
3 & 0.25 & 0.25 & 0.5 & 0.25 & 54.8 & 40.5 & 62.4 & 61.0 & 39.3 & 50.8 & 58.6 & 52.4 & 51.6 & 50.0 \\
4 & 0.5 & 1.0 & 0.5 & 0.25 & 55.6 & 40.5 & 61.7 & 60.7 & 39.6 & 50.6 & 58.7 & 52.8 & 51.0 & 50.0 \\
5 & 0.5 & 0.5 & 0.5 & 0.25 & 55.2 & 40.9 & 61.7 & 60.5 & 39.6 & 50.7 & 58.5 & 52.4 & 51.7 & 49.8 \\
6 & 0.5 & 0.25 & 1.0 & 0.25 & 55.1 & 40.5 & 62.4 & 60.9 & 39.4 & 50.9 & 58.6 & 52.2 & 51.0 & 49.4 \\
7 & 0.5 & 0.25 & 0.25 & 0.25 & 55.8 & 40.5 & 61.9 & 61.1 & 40.1 & 50.8 & 58.9 & 52.5 & 51.2 & 49.2 \\
8 & 0.5 & 0.25 & 0.5 & 1.0 & 55.5 & 40.3 & 61.8 & 60.8 & 39.9 & 50.6 & 58.7 & 52.9 & 51.8 & 50.1 \\
9 & 0.5 & 0.25 & 0.5 & 0.5 & 55.5 & 40.8 & 62.3 & 60.7 & 39.8 & 50.8 & 58.9 & 52.4 & 50.8 & 49.3 \\ \hline
\end{tabular*}

\vspace{-3mm}
\end{table*}

\subsubsection{\textbf{Effects of Trade-off Parameters}}

We further study the impact of the trade-off parameters $\beta_1$--$\beta_4$ in the overall loss function, which regulate the relative importance of segmentation supervision, context consistency, 2D keypoint alignment, and multi-view 3D consistency. As shown in Table~\ref{tab:tradeoff}, model performance remains stable when these parameters vary within a reasonable range, indicating that the FUSION framework is not overly sensitive to moderate changes in loss weighting. Among all tested configurations, Variation~1 provides the best overall results, reflecting a balanced contribution of the four auxiliary objectives and serving as our default setting.

\begin{table*}[t]
\centering
\caption{Ablation study on the Neighbor Relation branch design on RealMvMoAT sub-set.}
\vspace{-3mm}
\label{tab:neighbor_ablation}

\small
\renewcommand\arraystretch{0.90}
\setlength{\tabcolsep}{9.3pt}
\begin{tabular*}{\textwidth}{ l | c c | c c c | c c c c c}
    \hline
        \multirow{2}*{Variations} & \multicolumn{2}{c|}{Cross-view } & \multicolumn{3}{c|}{Temporal} & \multicolumn{5}{c}{Overall} \\ \cline{2-11}
         & AF1 & MHAA & IDF1 & MOTA & HOTA & $\mathcal{A}$ & $\mathcal{F}$ & $\mathcal{S}$@5 & $\mathcal{S}$@10 & $\mathcal{S}$@30\\ \hline\hline
Full branch (Ours) & 55.6 & 40.7 & 62.0 & 60.9 & 39.8 & 50.8 & 58.8 & 52.7 & 51.3 & 49.6 \\
w/o distance calibration & 53.9 & 39.2 & 60.1 & 59.0 & 37.7 & 49.1 & 57.0 & 50.3 & 49.2 & 48.2 \\
w/o GATv2 (Avg. Pooling) & 54.0 & 39.6 & 59.9 & 59.2 & 37.5 & 49.4 & 57.0 & 50.6 & 49.7 & 48.4 \\ 
\hline
\end{tabular*}

\vspace{-3mm}
\end{table*}

\subsubsection{\textbf{Effects of the Neighbor Relation Branch's Designs}}
We further evaluate two key designs in the neighbor relation branch: height-based distance calibration and GATv2-based relational modeling. 
Results are reported in Table~\ref{tab:neighbor_ablation}.
(1) Removing height correction (``w/o distance calibration'') measures neighbor distances only in 2D pixel space and ignores depth cues in ground-view scenes, which degrades association accuracy. This suggests that the proposed calibration better reflects realistic geometric relations among pedestrians.
(2) Replacing GATv2 with simple distance-weighted average pooling (``w/o GATv2 (Avg. Pooling)'') also reduces performance. Unlike average pooling, which treats neighbors symmetrically, GATv2 adaptively weights neighbor interactions and better distinguishes informative neighbors from noisy ones, improving robustness to occlusion and viewpoint changes.
These results show that both distance calibration and graph modeling are important for learning discriminative neighbor-aware features.

\begin{table*}[t]
\centering
\caption{Complexity and latency comparison between FUSION and the baseline on RealMvMoAT.}
\vspace{-3mm}
\label{tab:efficiency}

\small
\renewcommand\arraystretch{0.90}
\setlength{\tabcolsep}{5.7pt}
\begin{tabular*}{\textwidth}{ l | c c c | c c c c }
    \hline
        Method 
        & Params (M) 
        & FLOPs (G) 
        & Memory (MB)
        & Feature Extract (s) 
        & Cross-view Assoc. (s) 
        & Tracker (s) 
        & Total (s) \\ \hline\hline

Baseline & 21.67 & 4.91 & 694 & 51.36 & 2.59 & 75.98 & 133.48 \\
MvMHAT* & 23.16 & 5.12 & 790 & 51.22 & 3.23 & 51.89 & 109.84 \\
FUSION (Ours) & 26.35 & 5.50 & 998 & 75.56 & 3.35 & 56.96 & 135.87 \\ \hline

\end{tabular*}
\vspace{-3mm}
\end{table*}

\begin{figure}[t]
    \centering
    \includegraphics[width=0.95\linewidth]{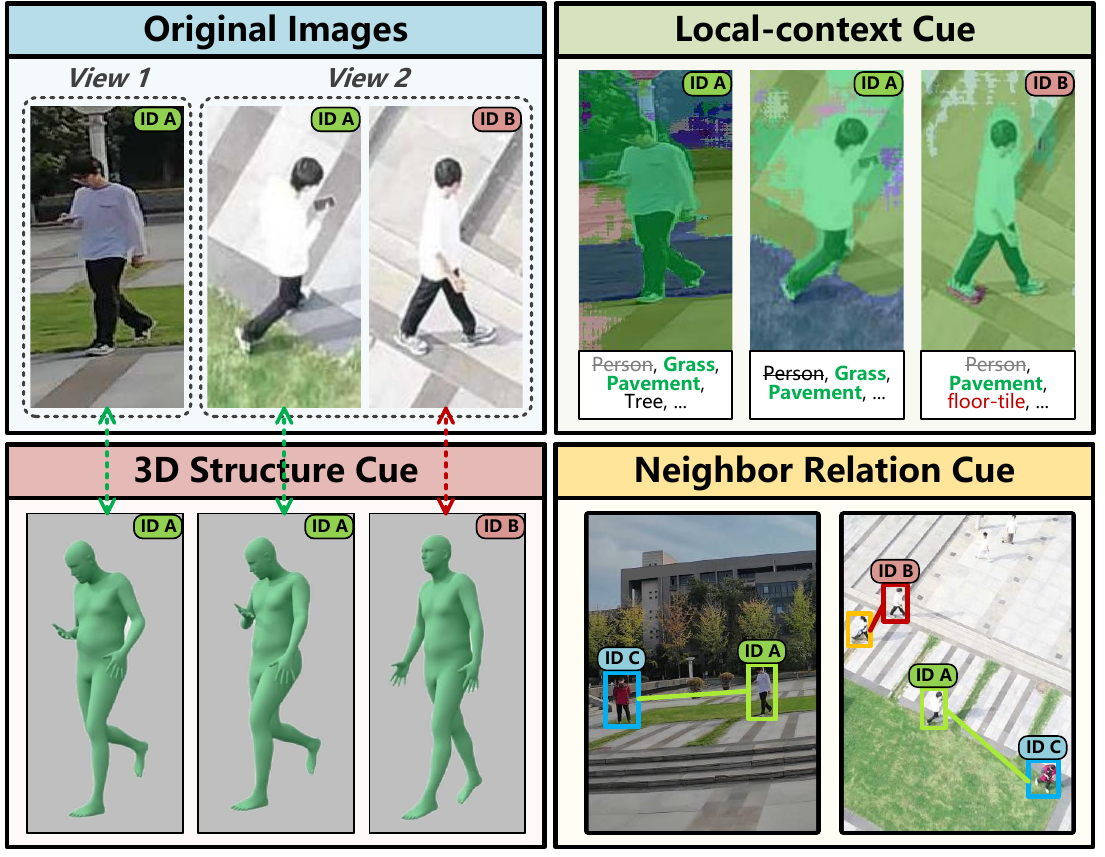}
    \vspace{-3mm}
    \caption{Visualization of a challenging case where two pedestrians exhibit highly similar appearances under different viewpoints.
    While the appearance alone leads to confusion between ID A and ID B, the additional cues provide discriminative context semantics, consistent 3D body structure, and stable spatial relations, allowing the model to correctly associate ID A across views. }
    \label{fig:vis}
    \vspace{-4mm}
\end{figure}

\begin{figure}[htbp] 
	\centering
	% \vspace{-2.5mm}
 \includegraphics[width=0.95\linewidth,scale=0.95]{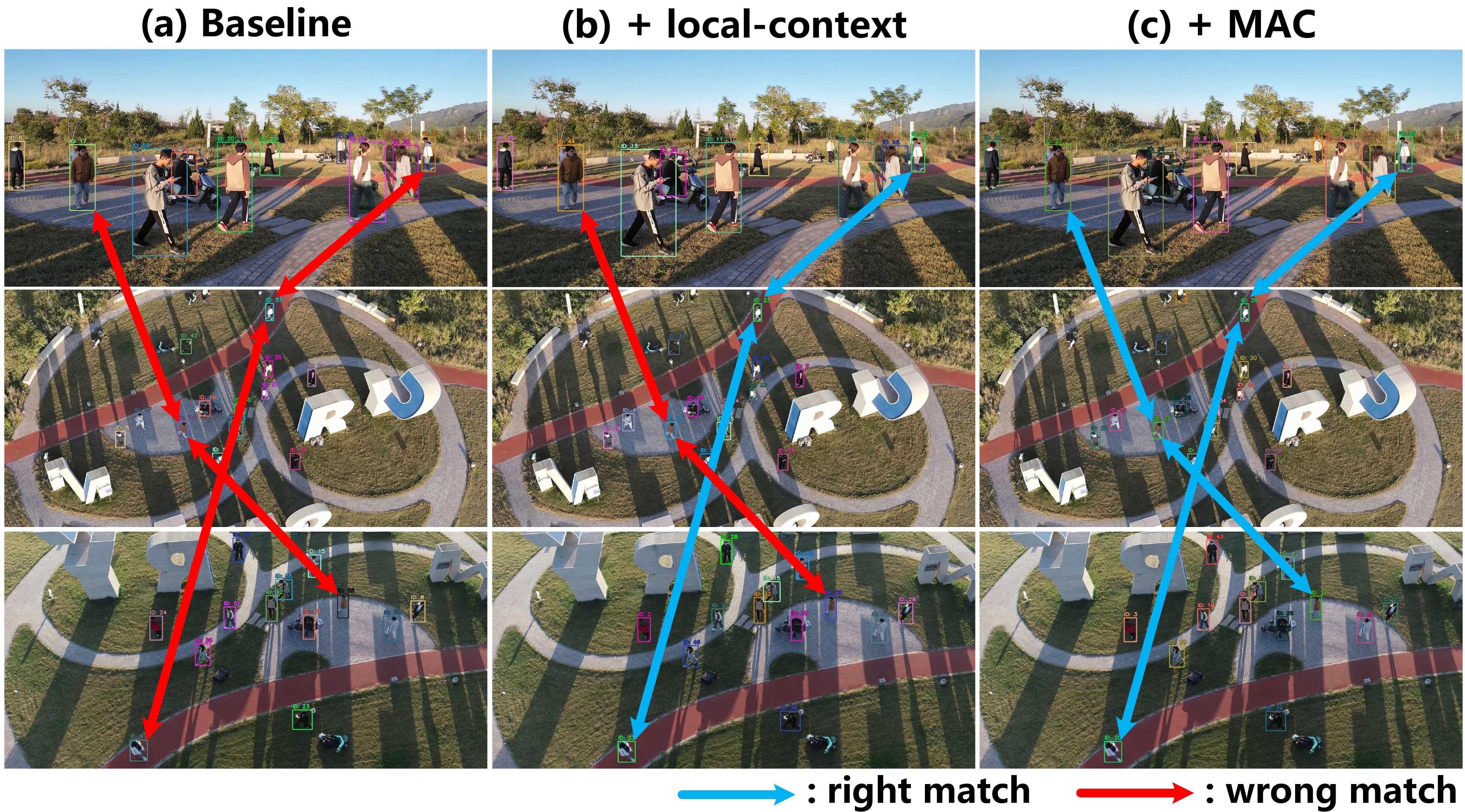}
 \vspace{-3mm}
	\caption{Qualitative visualization of cross-view matching results. Rows represent camera views, and columns represent different variations, and blue and red arrows represent correctly matched pairs and mismatched pairs.
} 
	\label{fig:datasets_base-cue}
\vspace{-4mm}
\end{figure}

\subsubsection{\textbf{Efficiency and Latency Analysis}}
We compare the computational efficiency of FUSION with the baseline. 
Table~\ref{tab:efficiency} reports model parameters, FLOPs, memory usage, and the latency of each sub-system, including feature extraction, cross-view association, and tracking, where latency is measured on a 500-frame, 5-camera sequence.
Although FUSION introduces additional branches and an adaptive fusion unit, the increases in parameters and FLOPs are modest, keeping the overall computational cost close to that of the baseline.
In latency, feature extraction is slower because the multi-branch architecture processes extra cue-specific heads. 
However, FUSION generates cleaner and more stable identity trajectories, resulting in fewer redundant or fragmented tracklets and thus a more efficient temporal tracking stage with fewer association operations. 
This partly offsets the added feature extraction cost and keeps the overall latency competitive.
Overall, FUSION achieves a favorable balance between accuracy and efficiency, significantly improving cross-view and temporal association performance with only small and practically acceptable overhead.

\subsubsection{\textbf{Visualization of Adaptive Fusion Weights Across Time and Scenes}}
We analyze the adaptive fusion mechanism in our FUSION framework by tracking how cue weights evolve across time and scenes. As shown in Fig. 6 (in Supplementary Materials), these weights adjust dynamically based on scene complexity and temporal variations.

\subsubsection{\textbf{Visualization of auxiliary cues}}
In Fig.~\ref{fig:vis}, we visualize a representative case to illustrate how auxiliary cues complement appearance features for cross-view association. Two pedestrians (ID~A and ID~B) share very similar clothing, making appearance alone unreliable, yet the auxiliary cues provide clear distinctions. 
\textbf{The local-context cue} shows that images of ID~A captured from different viewpoints consistently contain similar background semantics, \ie, \emph{grass} and \emph{pavement}.
In contrast, image of ID~B only contains \emph{pavement}, resulting in a significant semantic difference in environmental information from ID~A, which facilitates identity discrimination.
\textbf{The 3D structure cue} further reinforces this distinction: the reconstructed 3D meshes of ID~A are visually consistent across views, while those of ID~B exhibit noticeable discrepancies, indicating that the 3D representation captures intrinsic geometric properties that remain stable under perspective changes. Additionally, \textbf{the neighbor-relation cue} indicates that ID~A (green box) remains spatially close to ID~C (blue box) across all views, whereas ID~B (red box) is consistently distant from ID~C, further aiding accurate identity discrimination. 

\subsubsection{\textbf{Qualitative example of cross-view matching performance}}
Fig.~\ref{fig:datasets_base-cue} illustrates how our FUSION framework improves cross-view matching by leveraging auxiliary cues. In Fig.~\ref{fig:datasets_base-cue}(a), the baseline approach struggles with large viewpoint shifts, causing mismatches due to appearance distortions. Fig.~\ref{fig:datasets_base-cue}(b) demonstrates how the local-context cue helps mitigate this by stabilizing matching under environmental noise, such as occlusions and background changes. Finally, in Fig.~\ref{fig:datasets_base-cue}(c), the adaptive fusion mechanism further enhances stability by adjusting the weights of the cues. By giving more importance to reliable cues and reducing the influence of less reliable ones, the system achieves more accurate identity matching, even under challenging cross-view conditions.
\section{Conclusion}

In this work, we propose the FUSION framework for Multi-view Multi-object Association and Tracking, which integrates multiple auxiliary cues for robust cross-view matching and stable temporal tracking. Our approach, the FUSION framework, leverages the Multi-cue Adaptive Combination module and Online Multi-view Feature Synchronization to effectively address challenges like viewpoint shifts and occlusions.
We also introduce the RealMvMoAT dataset, a large-scale benchmark designed for real-world mobile multi-view perception. Extensive experiments show that FUSION outperforms existing methods, offering a strong foundation for future research in mobile perception systems.

% \section*{Acknowledgments}
% This should be a simple paragraph before the References to thank those individuals and institutions who have supported your work on this article.

% {\appendix[Proof of the Zonklar Equations]
% Use $\backslash${\tt{appendix}} if you have a single appendix:
% Do not use $\backslash${\tt{section}} anymore after $\backslash${\tt{appendix}}, only $\backslash${\tt{section*}}.
% If you have multiple appendixes use $\backslash${\tt{appendices}} then use $\backslash${\tt{section}} to start each appendix.
% You must declare a $\backslash${\tt{section}} before using any $\backslash${\tt{subsection}} or using $\backslash${\tt{label}} ($\backslash${\tt{appendices}} by itself
%  starts a section numbered zero.)}


\bibliographystyle{IEEEtran}
\bibliography{main}
% \end{thebibliography}

\vfill

\end{document}